\documentclass[11pt]{article}

\usepackage[final]{acl}
\usepackage{times}
\usepackage{latexsym}
\usepackage[T1]{fontenc}
\usepackage[utf8]{inputenc}
\usepackage{microtype}
\usepackage{inconsolata}
\usepackage{graphicx}
\usepackage{amsmath}
\usepackage{amssymb}
\usepackage{booktabs}
\usepackage{xcolor}
\usepackage{tcolorbox}
\usepackage{array}

\title{QAQ: Bidirectional Semantic Coherence for Selecting High-Quality Synthetic Code Instructions}
\author{Jiayin Lei$^{1}$\thanks{Equal contribution. $^{\dagger}$Corresponding author.}, \ 
        Ming Ma$^{2,3\ast}$, \ 
        Yunxi Duan$^{1}$, \ 
        Chenxi Li$^{4}$, \ 
        Tianming Yang$^{2\dagger}$ \\
        $^{1}$School of Mathematics, Statistics and Mechanics, Beijing University of Technology, Beijing, 100124, China \\
        $^{2}$Institute of Neuroscience, State Key Laboratory of Brain Cognition and Brain-inspired \\
        Intelligence Technology, Center for Excellence in Brain Science and Intelligence Technology, \\
        Chinese Academy of Sciences, Shanghai, 200031, China \\
        $^{3}$School of Future Technology, University of Chinese Academy of Sciences, Beijing, 100049, China \\
        $^{4}$University of Chicago, Chicago, IL, USA \\
        \texttt{raylanxxsg559@gmail.com, \{mam2022, tyang\}@ion.ac.cn}
}
\begin{document}
\maketitle

\begin{abstract}
    
Synthetic data has become essential for training code generation models, yet it introduces significant noise and hallucinations that are difficult to detect with current metrics. Existing data selection methods like Instruction-Following Difficulty (IFD) typically assess how hard a model generates an answer given a query ($A|Q$). However, this metric is ambiguous on noisy synthetic data, where low probability can distinguish between intrinsic task complexity and model-generated hallucinations. Here, we propose QAQ, a novel data selection framework that evaluates data quality from the reverse direction: how well can the answer predict the query ($Q|A$)? We define Reverse Mutual Information (RMI) to quantify the information gain about the query conditioned on the answer. Our analyses reveal that both extremes of RMI signal quality issues: low RMI indicates semantic misalignment, while excessively high RMI 
may contain defect patterns that LLMs easily recognize. 
Furthermore, we introduce a selection strategy based on the disagreement between strong and weak models to identify samples that are valid yet challenging. Experiments across three datasets spanning code generation (WarriorCoder, Magpie-Qwen2.5-Coder-Pro-300K) and math reasoning (OpenR1-Math-220k) demonstrate that selecting just 25\% of data using stratified RMI matches full-data performance while being consistently competitive with or better than existing data selection methods. Our approach highlights the importance of bidirectional semantic coherence in synthetic data curation, offering a scalable pathway to reduce computational costs without sacrificing model capability. Code is available at \url{https://github.com/XXSg559/QAQ}.
\end{abstract}

\begin{figure}[t] 
\centering
\includegraphics[width=0.6\columnwidth]{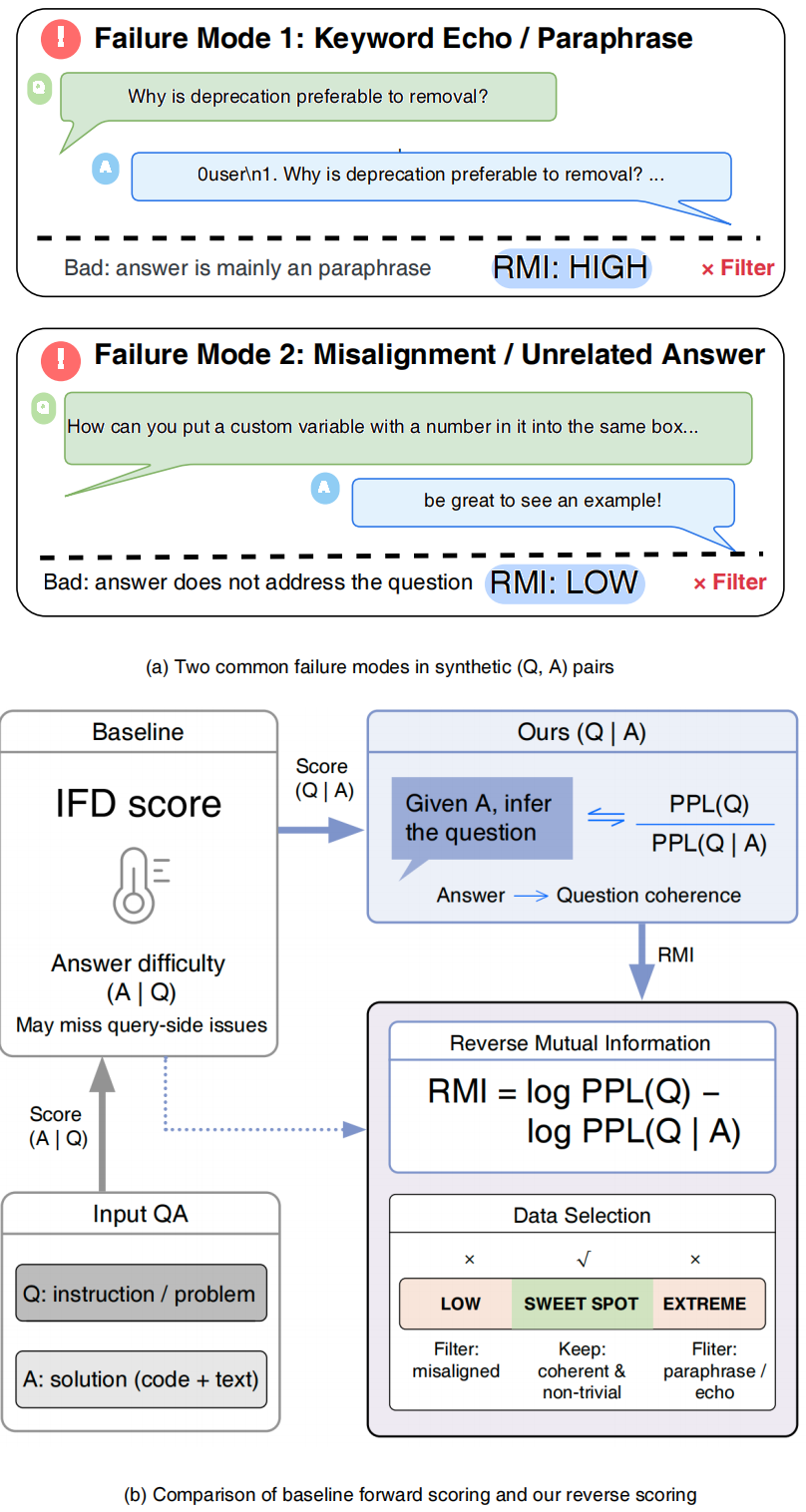}
\caption{\textbf{(a)} Two common failure modes in synthetic $(Q,A)$: keyword echo/paraphrase (high RMI) and misalignment (low RMI). \textbf{(b)}  The comparisons between our method and
traditional IFD method. Unlike previous work, QAQ scores question predictability given the answer $(Q | A)$ and computes RMI.
% $\mathrm{RMI}=\log\mathrm{PPL}(Q)-\log\mathrm{PPL}(Q | A)$. 
We keep the mid-range ``sweet spot'' and filter both extremes.}

\label{fig:fig1a}
\end{figure}
\section{Introduction}

Synthetic data has become the dominant paradigm for training code generation models \citep{chen2021codex,codellama,li2023starcoder}. Seedless data synthesis methods like Magpie \citep{magpie} and WarriorCoder \citep{warriorcoder} generate diverse instruction-response pairs by prompting aligned LLMs directly, without relying on human-curated seed data. While this approach enables broad diversity, it also produces massive amounts of noisy data that are difficult to curate.

The quality issues in synthetic data are subtle and often undetectable by surface-level metrics. We observe two failure modes in practice: (1) queries containing fabricated terminology (e.g., ``convert a gritty sulla matrix''), where models generate syntactically valid code that echoes these made-up terms, producing high surface-level quality scores while providing little learning signal; (2) off-topic queries that should not appear in a code dataset at all (e.g., a greeting ``Hi!''), paired with incoherent responses (e.g., starting with ``compensated...'')---indicating a completely broken Q-A pair. These failure modes are difficult for answer-focused metrics to detect.

Notably, methods like WarriorCoder focus on \textbf{answer quality}: multiple models generate competing solutions, and the highest-scored answer is selected for each query. However, this pipeline largely overlooks \textbf{query quality}.

Existing data selection methods like Instruction-Following Difficulty (IFD) \citep{cherry_llm} operate in the $A|Q$ direction.
% , measuring how hard it is to generate the answer given the query. This captures generation difficulty rather than query-answer coherence. 
Although mutual information is theoretically symmetric, due to the autoregressive nature of language models, its practical estimation via conditional perplexity is not necessarily consistent. Empirically, we find low correlation ($\rho = 0.252$) between the two, confirming they capture distinct aspects of data quality.

We propose \textbf{QAQ} (Query, Answer-Query), a method that measures quality from the reverse direction: $\text{PPL}(Q|A)$, which captures how well the answer \textit{explains} the question. The intuition is analogous to a student who struggles to understand a problem, but finds clarity after seeing the solution. We define RMI as $\log \text{PPL}(Q) - \log \text{PPL}(Q|A)$, which captures the information gain that the answer provides about the question:
\begin{itemize}
    \item \textbf{Low RMI}: The answer provides little information gain about the question; seeing the answer does not help predict the question.
    \item \textbf{Very high RMI}: 
    The answer makes the query trivially predictable, often due to superficial or shortcut patterns such as paraphrasing or repetition.
    % often indicating defect patterns that LLMs easily recognize, such as paraphrasing or repetition.
\end{itemize}

Furthermore, we observe that strong and weak models disagree on sample quality. Samples where only the strong model recognizes high RMI may contain sophisticated patterns that weak models cannot capture. We call this the ``cognitive gap''.

To test whether these ideas generalize beyond code, we additionally transfer QAQ to mathematical reasoning.

Our contributions are:
\begin{itemize}
    \item We identify that seedless synthetic data pipelines overlook query quality, and propose QAQ, a data selection method based on Reverse Mutual Information (RMI) that captures query-answer semantic alignment.
    \item 
    We empirically show that RMI serves as an effective quality indicator, where both misaligned and shortcut-dominated samples can be filtered through its value distribution.
    % We show that both extremes of RMI indicate quality problems: low RMI suggests misalignment, while extremely high RMI may contain defect patterns that LLMs easily recognize.
    \item We introduce model disagreement as a selection signal, capturing samples with high ``cognitive gap'' between strong and weak models.
    \item We demonstrate that 25\% of data selected by our method matches full-data performance while being consistently competitive with or better than existing data selection methods across both code generation and math reasoning domains.
\end{itemize}

\begin{figure*}[t] 
    \centering
    \includegraphics[width=0.9\textwidth]{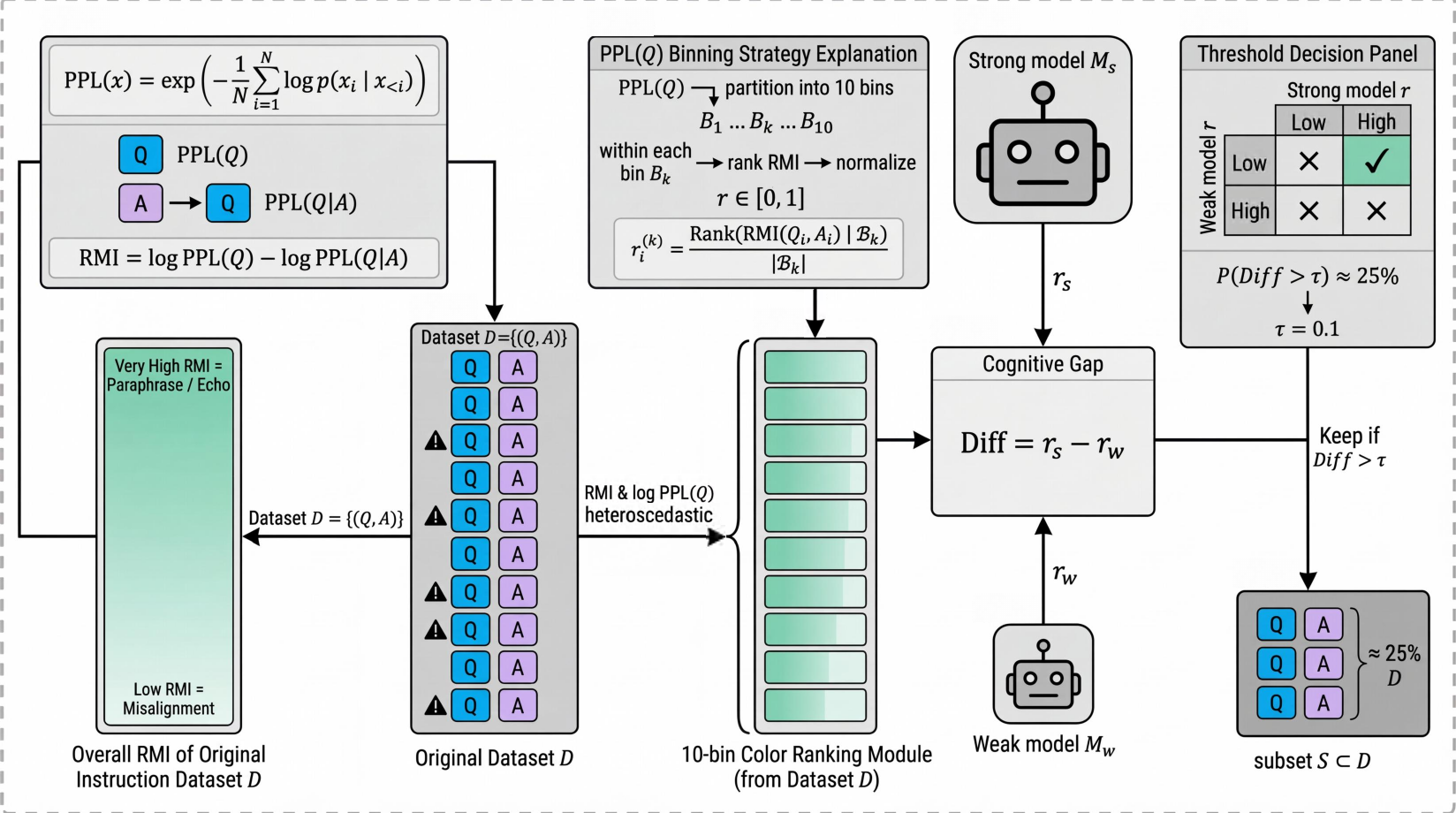} 
    \caption{Overall workflow of the QAQ framework for synthetic code data selection. 
    \textbf{(Top-left)} RMI is calculated as the information gain provided by the answer about the query, using reverse perplexity. 
    \textbf{(Bottom-left)} Our analysis identifies that RMI extremes signal quality issues: low RMI indicates semantic misalignment, while excessively high RMI indicates trivial paraphrasing or echoing. 
    \textbf{(Center)} To eliminate the confounding factor of question complexity, RMI is stratified into 10 bins based on $\text{PPL}(Q)$ to compute normalized within-bin ranks $r$. 
    \textbf{(Right)} The final selection targets the ``Cognitive Gap'' ($\text{Diff} = r_s - r_w$) between a strong model $M_s$ and a weak model $M_w$. Samples that are recognized as high-quality only by the strong model are retained, resulting in a high-quality 25\% subset $S$ that provides strong learning signals.}
    \label{fig:large_rmi_analysis}
\end{figure*}

\section{Related Work}

\paragraph{Data Selection for Instruction Tuning.}
Several approaches have been proposed to select high-quality samples from instruction-tuning datasets. IFD \citep{cherry_llm} measures sample quality using $\text{PPL}(A|Q)/\text{PPL}(A)$, the ratio of conditional to unconditional answer perplexity. Samples with IFD close to (but less than) 1 are preferred: IFD $\ll 1$ suggests the answer is trivially predictable from the instruction, while IFD $> 1$ indicates the instruction actually hinders answer generation. Superfiltering \citep{superfiltering} shows that weak models can approximate strong models' IFD scores, enabling efficient filtering at scale. These methods operate in the $A|Q$ direction, measuring generation difficulty rather than query-answer semantic alignment.

Beyond perplexity-based methods, gradient-based approaches like LESS \citep{less} use influence estimation to identify training samples most relevant to target tasks. Multi-dimensional methods like D3 \citep{d3} jointly optimize diversity, difficulty, and dependability via coreset objectives, while RDS+ \citep{rdsplus} selects samples by cosine similarity between pretrained hidden-state representations and a small set of task-specific seed examples. Recent work explores orthogonal selection criteria: T-SHIRT \citep{tshirt} improves IFD through token-level selection and neighborhood robustness; SCAR \citep{scar} targets style consistency between instructions and responses; ProDS \citep{prods} aligns selection with task-specific preferences via preference optimization. These methods optimize for diversity, style, or preference alignment, but do not directly assess query-answer semantic coherence.

\paragraph{Synthetic Data for Code Generation.}
High-quality datasets are essential for code generation. Code Alpaca \citep{codealpaca} and WizardCoder \citep{wizardcoder} use Evol-Instruct to evolve seed instructions into more complex programming problems. OSS-Instruct \citep{magicoder} generates diverse problems by extracting coding concepts from open-source code snippets. These seed-based methods inherit biases from their curated seeds, constraining diversity to the patterns present in the original data. In contrast, seedless approaches like Magpie \citep{magpie} and WarriorCoder \citep{warriorcoder} achieve greater diversity by prompting aligned LLMs directly without relying on human-curated data. However, this freedom comes at a cost: without seed constraints, the generated data contains more noise and semantic inconsistencies. While WarriorCoder selects the highest-scored answer for each query through cross-model evaluation, query quality itself remains largely unaddressed.

\paragraph{Mutual Information in NLP.}
Pointwise mutual information (PMI) has been widely used in NLP for measuring word associations and semantic similarity \citep{church1990pmi}. At the token level, PMI helps identify collocations and construct semantic spaces. This concept can be extended to the sequence level, where mutual information between query and answer sequences reflects their semantic coherence. A related reverse-direction idea appears in summarization evaluation: \citet{scialom2019answers} score a generated summary by how well it lets an automatic QA system answer questions derived from the source, using answerability as an unsupervised proxy for content overlap. We adopt a similar reverse-direction intuition, but apply it to conditional perplexity for training data selection rather than post-hoc evaluation.

\paragraph{Model Comparison for Data Quality.}
Using multiple models to assess data has proven effective in various contexts. \citet{helm2024} compare short-context and long-context models to weight tokens for long-range language modeling. Rho-1 \citep{rho1} compares trained and untrained versions of the same model to identify tokens with high ``excess loss'' for selective pre-training. These works focus on token-level selection, while sample-level selection using model disagreement remains underexplored.

\section{QAQ: Reverse Mutual Information for Data Selection}

Given a large-scale instruction-tuning dataset $\mathcal{D} = \{(Q_i, A_i)\}_{i=1}^{N}$ for code generation, our goal is to select a high-quality subset $\mathcal{S} \subset \mathcal{D}$ that achieves comparable performance to training on the full dataset while significantly reducing computational costs.

\subsection{Reverse Mutual Information}

Prior work such as IFD \citep{cherry_llm}
% measures data quality using $\text{PPL}(A|Q)/\text{PPL}(A)$, which 
captures whether the instruction facilitates answer generation. Samples with IFD close to (but less than) 1 are preferred, representing what the model ``does not yet know''.

We propose an alternative perspective: instead of measuring generation difficulty, we measure how well the answer \textit{explains} the question. Specifically, we compute the \textbf{Reverse Mutual Information (RMI)} as:
\begin{equation}
\text{RMI}(Q, A) = \log \text{PPL}(Q) - \log \text{PPL}(Q|A)
\end{equation}

where $\text{PPL}(Q)$ is the perplexity of the question alone, and $\text{PPL}(Q|A)$ is the perplexity of the question conditioned on the answer. Perplexity is computed as:
\begin{equation}
\text{PPL}(x) = \exp\left(-\frac{1}{N}\sum_{i=1}^{N}\log p(x_i \mid x_{<i})\right)
\end{equation}
where $N$ is the number of tokens in sequence $x$.

This formulation is inspired by the mutual information $I(Q; A) = H(Q) - H(Q|A)$, capturing an \textit{information-gain}-like quantity that reflects how much the answer helps predict the question under our scoring model. Intuitively:
\begin{itemize}
    \item High RMI indicates that seeing the answer makes the question much easier to predict, suggesting strong semantic coherence between Q and A.
    \item Low RMI suggests the answer does not relate well to the question (e.g., copy-paste errors, irrelevant responses).
\end{itemize}

\paragraph{Comparison with IFD.} While IFD operates in the $A|Q$ direction, measuring ``instruction-following difficulty'', our approach operates in the $Q|A$ direction, measuring ``answer explanatory power''. We argue that a good code solution should clearly reflect the original problem specification: given the code, one should be able to infer what problem it solves.

\begin{figure}[t]
\centering
\includegraphics[width=0.9\columnwidth]{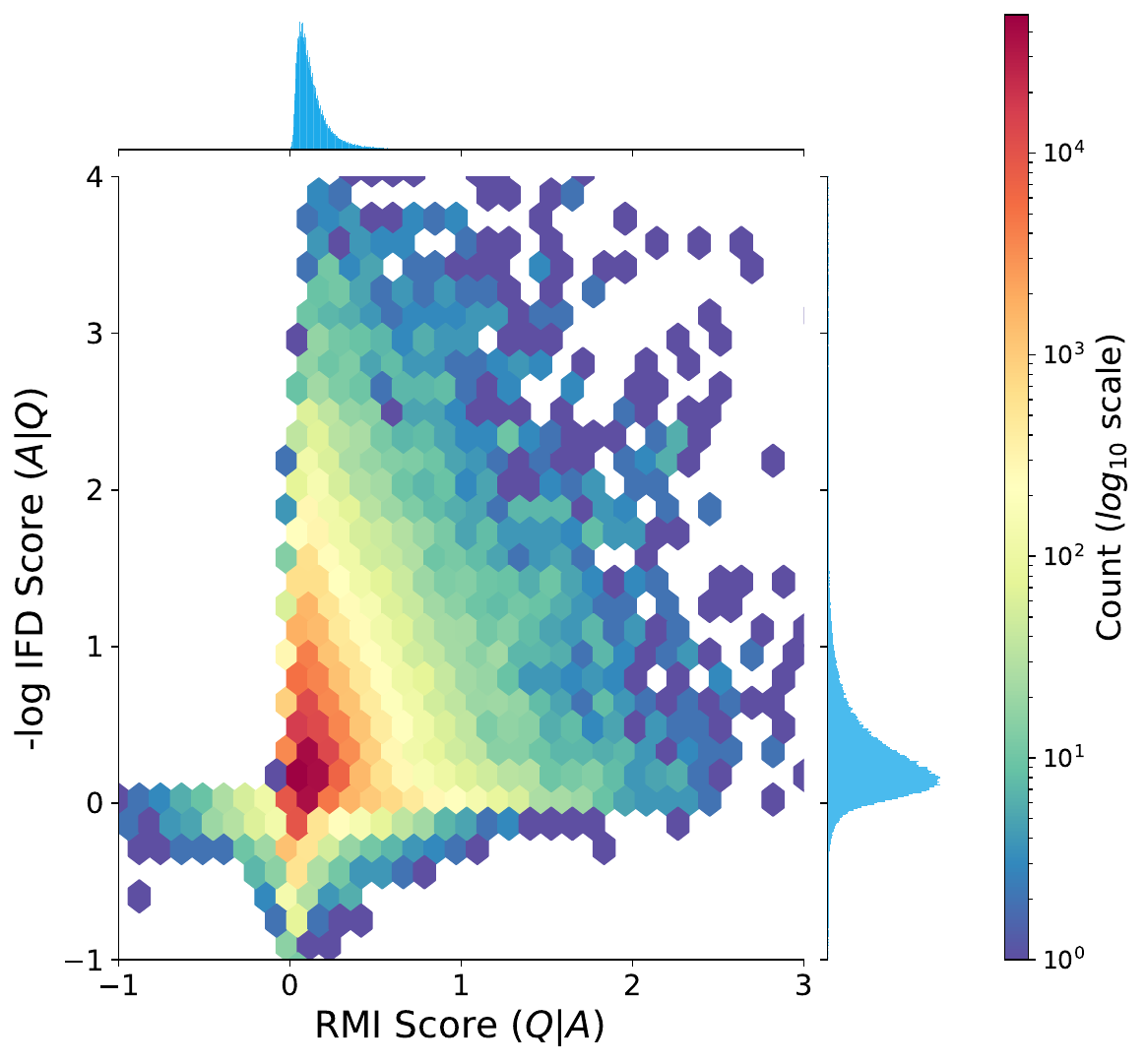}
\caption{Scatter plot of RMI (Question-based) vs. -log(IFD) (Answer-based). Despite both being mutual information estimates, the Spearman rank correlation is low ($\rho = 0.252$), suggesting they capture different aspects of data quality.}
\label{fig:rmi_ifd}
\end{figure}

\paragraph{Why the Two Directions Differ.} As noted in the introduction, the $A|Q$ and $Q|A$ directions capture different signals due to length disparity, information density differences, and model capability asymmetry. This is consistent with the view of predictive $\mathcal{V}$-information \citep{xu2020vinfo}, which shows that for a bounded predictor, conditioning in opposite directions generally yields different amounts of usable information, unlike symmetric Shannon mutual information. Figure~\ref{fig:rmi_ifd} shows the low correlation ($\rho = 0.252$) between RMI and IFD, confirming they measure distinct aspects of data quality.

\paragraph{Implementation of $\text{PPL}(Q)$.} We compute $\text{PPL}(Q)$ using the standard chat template, which matches the format used during training:

\begin{tcolorbox}[colback=blue!5!white, colframe=blue!50!white, boxrule=0.5pt]
\small
\textbf{System}: Standard system prompt \\
\textbf{User}: \texttt{[Q]}
\end{tcolorbox}

\noindent We compute $\text{PPL}(Q)$ on the question tokens in the user turn via teacher-forcing with per-token length normalization.

\paragraph{Implementation of $\text{PPL}(Q|A)$.} A naive implementation would simply concatenate the answer before the question, but this creates an unnatural input order. Instead, we frame the computation as a \textit{reverse generation task} using the chat template:

\begin{tcolorbox}[colback=blue!5!white, colframe=blue!50!white, boxrule=0.5pt]
\small
\textbf{System}: Standard system prompt \\
\textbf{User}: \texttt{TASK: Given an answer, generate the most likely computer science question that this answer is responding to. If the inferred question is outside computer science, respond with ``INVALID''. Answer: [A]} \\
\textbf{Assistant}: \texttt{[Q]}
\end{tcolorbox}

\noindent We compute $\text{PPL}(Q|A)$ via teacher-forcing on the ground-truth question tokens in the assistant turn, with per-token length normalization. While the two computations use different message roles (user vs. assistant), the RMI score captures how much the answer $A$ reduces uncertainty about $Q$. The relative difference is meaningful even across different positional contexts. Full prompt templates and system configurations are provided in Appendix~\ref{sec:appendix_prompts}.

\paragraph{Validation of Chat Template.} 
We verify that IFD remains highly robust across template variations ($\rho=0.957$). In contrast, RMI is sensitive to context: removing the template causes a ``cold start'' in $PPL(Q)$ calculation, where the lack of context artificially inflates perplexity and biases the score. This suggests that template context is important for RMI to capture genuine semantic coherence. We provide a detailed sensitivity analysis in Appendix~\ref{sec:appendix_template}.

\subsection{Stratified RMI}

\begin{figure}[t]
\centering
\includegraphics[width=0.9\columnwidth]{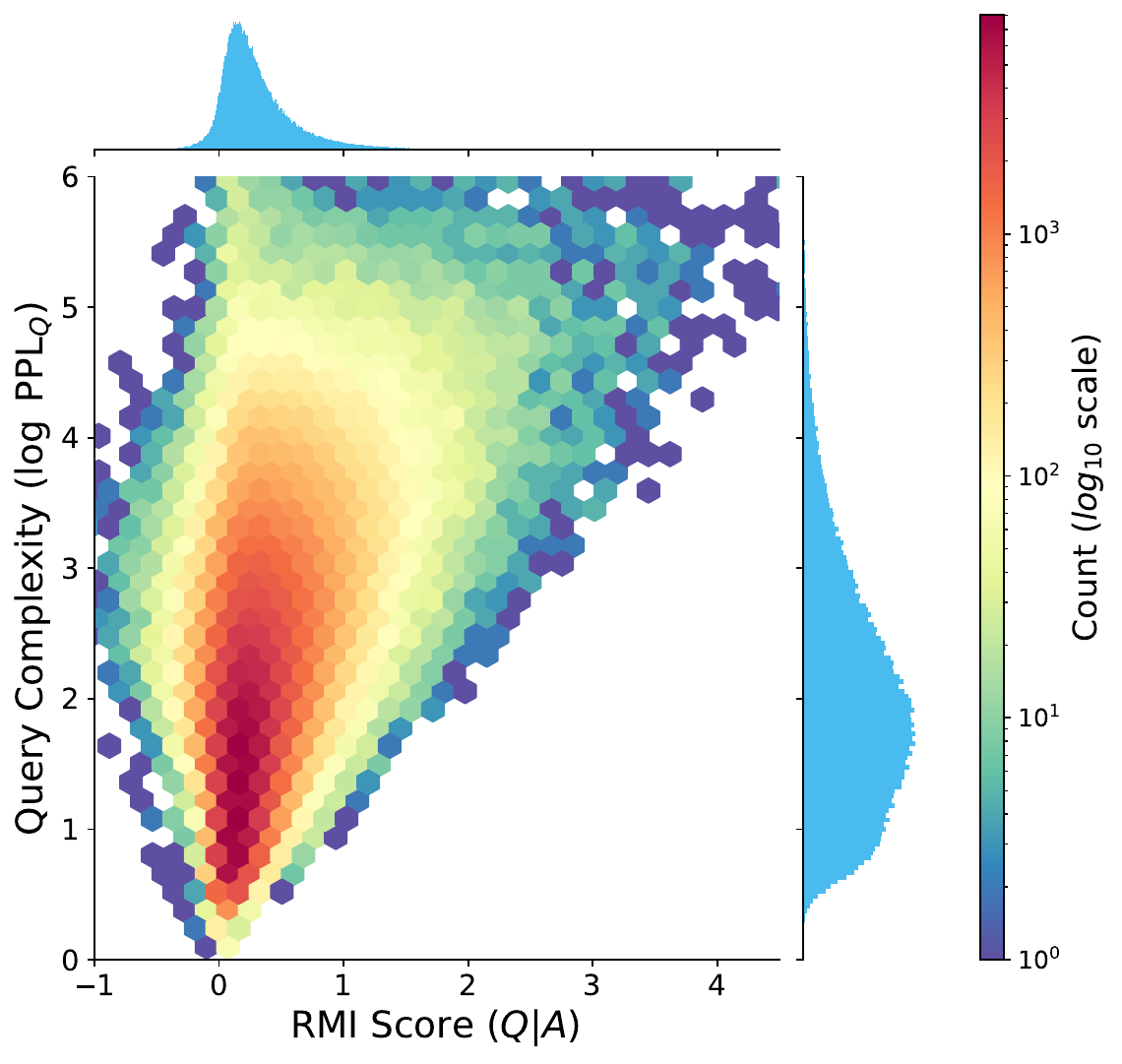}
\caption{Scatter plot of RMI vs. $\log \text{PPL}(Q)$. The heteroscedastic pattern (low variance at low PPL, high variance at high PPL) motivates stratification by question complexity.}
\label{fig:rmi_ppl}
\end{figure}

Directly filtering by global RMI rankings introduces a confounding factor: question complexity. Simple questions
% (e.g., ``print hello world'') 
naturally have low $\text{PPL}(Q)$, leading to lower RMI scores regardless of answer quality. As shown in Figure~\ref{fig:rmi_ppl}, the relationship between RMI and $\log \text{PPL}(Q)$ exhibits clear heteroscedasticity: samples with low $\text{PPL}(Q)$ have narrowly distributed RMI values (concentrated at lower ranges), while samples with high $\text{PPL}(Q)$ show much greater RMI variance. This indicates that RMI and question complexity capture different aspects of data quality. Without stratification, simple but valid questions may be incorrectly penalized due to their inherently lower RMI ceiling.

To address this, we propose \textbf{Stratified RMI}: we first partition the dataset into $K=10$ bins (deciles) based on $\text{PPL}(Q)$, then compute RMI rankings independently within each bin. We use $K=10$ as a default, which our ablation study (Section~\ref{sec:analysis}) shows to be a stable choice across a range of values. This ensures that samples are compared against others of similar question complexity.

Formally, let $\mathcal{B}_k$ denote the $k$-th bin containing samples with similar $\text{PPL}(Q)$ values. For each sample $(Q_i, A_i) \in \mathcal{B}_k$, we compute its within-bin rank and normalize to $[0, 1]$:
\begin{equation}
r_i^{(k)} = \frac{\text{Rank}(\text{RMI}(Q_i, A_i) \mid \mathcal{B}_k)}{|\mathcal{B}_k|}
\end{equation}
where $|\mathcal{B}_k|$ is the number of samples in bin $k$.

\subsection{Model Disagreement}

As we show empirically in Section~\ref{sec:analysis}, mid-range RMI outperforms both extremes, yet manually specifying such percentile boundaries is arbitrary and dataset-dependent. We therefore seek a more principled selection criterion. Following prior work showing that weak models can approximate strong models' data quality scores \citep{superfiltering}, we first evaluate whether RMI exhibits similar model-invariance. While strong and weak models show high rank correlation (Spearman $\rho \approx 0.7$) and 76.1\% overlap in top-50\% selection, their downstream performance differs substantially (validated in Section~\ref{sec:analysis}). This suggests the non-overlapping samples carry disproportionate training signal.

We therefore explore \textbf{model disagreement} as an alternative signal. We compute stratified RMI ranks using two models:
\begin{itemize}
    \item A strong model $M_s$ (e.g., DeepSeek-Coder-6.7B-Base)
    \item A weak model $M_w$ (e.g., Qwen3-0.6B)
\end{itemize}

For each sample, we obtain two normalized ranks $r_s, r_w \in [0, 1]$ and define:
\begin{align}
\text{Diff} &= r_s - r_w \quad \text{(disagreement signal)}
\end{align}

High Diff indicates samples where the strong model gives high RMI rank while the weak model gives low RMI rank. We hypothesize that such samples are both \textit{valid} (recognized by strong model) and \textit{learnable} (challenging for weak model). We validate this hypothesis empirically in Section~\ref{sec:analysis}. Our final selection retains the top 25\% of samples ranked by Diff score. We observe that this cutoff falls at approximately $\text{Diff} > 0.1$ on WarriorCoder.

\section{Experiments}

\subsection{Experimental Setup}

\paragraph{Dataset.} We use a reproduction of the WarriorCoder dataset \citep{warriorcoder} from HuggingFace, containing 329K instruction-response pairs for code generation created through multi-agent collaboration. After filtering samples exceeding 2048 tokens, approximately 310K samples remain. The original dataset is not publicly available.

\paragraph{Models.} For computing RMI scores, we use:
\begin{itemize}
    \item Strong model: DeepSeek-Coder-6.7B-Base \citep{deepseek-coder}
    \item Weak model: Qwen3-0.6B \citep{qwen3}
\end{itemize}
For fine-tuning evaluation, we train DeepSeek-Coder-6.7B-Base on selected subsets.

\paragraph{Training Details.} We fine-tune for 3 epochs using LlamaFactory \citep{zheng2024llamafactory}. Due to the noisy nature of our dataset, we refrain from excessively reducing the batch size. Batch sizes and learning rates are adjusted by dataset size: for full data, we use batch size 512 and learning rate 1.2e-4; for 50\% data, batch size 256 and learning rate 0.8e-4; for 25\% data, batch size 256 and learning rate 0.4e-4. Additionally, the learning rate scheduler follows a cosine decay with a warmup ratio of 0.2.

\paragraph{Evaluation.} We evaluate on HumanEval \citep{chen2021codex}, HumanEval+, MBPP \citep{mbpp}, and MBPP+ benchmarks \citep{evalplus} using greedy decoding.

\paragraph{Comparisons.} We include two groups. \textbf{RMI ablation variants} vary the selection range and model choice:
\begin{itemize}
    \item RMI Top 50\%: Top 50\% by strong model (DeepSeek-Coder-6.7B-Base) stratified RMI
    \item RMI QW 50\%: Top 50\% by weak model (Qwen3-0.6B) stratified RMI
    \item RMI 50-75\%: Middle range by strong model stratified RMI
    \item RMI Top/Bottom-25\%: Extreme ranges by strong model stratified RMI
\end{itemize}
\textbf{External baselines} represent competing data selection methods, spanning perplexity-based, representation-based, and style-based selection mechanisms:
\begin{itemize}
    \item Random: Uniformly sample 25\% of the data without replacement.
    \item IFD \citep{cherry_llm}: Computes $\text{PPL}(A|Q)/\text{PPL}(A)$ using DeepSeek-Coder-6.7B-Base and retains the top 25\% of samples with IFD score closest to (but below) 1.
    \item RDS+ \citep{rdsplus}: Each $(Q, A)$ pair is embedded using the base model's own hidden states via a position-weighted mean pool of the last hidden layer (later tokens weighted more heavily), and cosine similarity is computed against a small set of task-specific seed examples (25 from HumanEval, 25 from MBPP). Samples are selected round-robin across the seed examples, taking the nearest unselected neighbor for each seed in turn, until the 25\% budget is reached.
    \item SCAR \citep{scar}: A trained style-consistency ranker (\texttt{lizhuang144/scar-gte-large}) scores each $(Q, A)$ pair on two axes: linguistic form (surface structure of the response, e.g.\ sentence structure and punctuation, independent of the instruction) and instructional surprisal (how closely $\text{PPL}(A|Q)$ matches human-written response patterns). The top 25\% by combined score are retained.
\end{itemize}

\noindent All perplexity computations use the same chat template format for fair comparison.

\paragraph{Math Domain.}\label{sec:math} To verify generalization beyond code, we apply QAQ to OpenR1-Math-220k \citep{openr1} (91K samples after filtering to 16,384-token cutoff), using Qwen2.5-Math-7B-Instruct \citep{qwen25math} as the base model and following OpenR1-Qwen-7B's training configuration \citep{openr1}. All methods select 25\% of data; QAQ uses Qwen2.5-Math-7B and Qwen2.5-Math-1.5B \citep{qwen25math} as strong and weak models with the same 10-bin stratification. We evaluate on MATH-500 \citep{lightman2023verify,hendrycks2021math} and GPQA-Diamond \citep{rein2023gpqa} (Pass@1) using the LightEval framework \citep{lighteval}.

\subsection{Main Results}

We fine-tune DeepSeek-Coder-6.7B-Base on selected subsets from WarriorCoder (310K samples) and evaluate on HumanEval(+) and MBPP(+) \citep{evalplus}. Results are shown in Table~\ref{tab:main_results}.

\begin{table}[t]
\centering
\scriptsize
\setlength{\tabcolsep}{2pt}
\begin{tabular}{lrcccc}
\toprule
\textbf{Method} & \textbf{\%} & \textbf{HumanEval} & \textbf{HumanEval+} & \textbf{MBPP} & \textbf{MBPP+} \\
\midrule
Full Data & 100 & 78.05 & 72.56 & 71.69 & 59.52 \\
\midrule
RMI Top 50\% & 50 & \textbf{78.05} & \textbf{73.17} & 72.22 & 58.20 \\
RMI 50\% (G) & 50 & 76.22 & 70.12 & \textbf{73.02} & \textbf{60.58} \\
RMI QW 50\% & 50 & 72.56 & 68.90 & 70.63 & 58.47 \\
\midrule
RMI 50-75\% & 25 & 77.44 & 72.56 & 71.43 & 58.47 \\
RMI Top-25\% & 25 & 73.17 & 70.12 & 70.11 & 57.14 \\
RMI Bot-25\% & 25 & 72.56 & 67.68 & 65.34 & 54.76 \\
\midrule
Random & 25 & 73.78 & 69.51 & 68.52 & 57.67 \\
IFD & 25 & 71.95 & 66.46 & 64.81 & 54.76 \\
RDS+ & 25 & 76.83 & 71.34 & 71.69 & 58.99 \\
SCAR & 25 & 75.00 & 70.73 & 70.63 & 57.67 \\
\midrule
\textbf{QAQ (Ours)} & 25 & 77.44 & 71.95 & 71.43 & 58.73 \\
\bottomrule
\end{tabular}
\caption{Main results (pass@1 with greedy decoding). (G) = global ranking. \textbf{QAQ (Ours)} selects the top 25\% of samples by Diff score ($\text{Diff} = r_s - r_w$); see Section~\ref{sec:analysis}.}
\label{tab:main_results}
\end{table}

Key findings:

\begin{itemize}
    \item \textbf{RMI outperforms most baselines}: At 25\% data, RMI 50-75\% outperforms Random, IFD, and SCAR on both benchmarks. It achieves results comparable to RDS+, surpassing it on HumanEval+ (72.56 vs 71.34) while remaining competitive on MBPP+.
    \item \textbf{Middle range is optimal}: RMI 50-75\% (25\% data) matches full-data performance on HumanEval+ (72.56) and approaches it on MBPP+ (58.47 vs 59.52). This ``sweet spot'' contains samples with good semantic alignment without trivial paraphrasing.
    \item \textbf{Both extremes are harmful}: RMI Top-25\% underperforms full data by 2.4/2.4 points on HumanEval+/MBPP+ (70.12/57.14 vs 72.56/59.52), suggesting extremely high RMI indicates certain defect patterns that language models easily recognize, such as answer paraphrasing the query. RMI Bottom-25\% underperforms by 4.9/4.8 points (67.68/54.76), confirming low RMI indicates semantic misalignment.
    \item \textbf{Model disagreement further improves HumanEval}: QAQ (25\% data, selected by model disagreement) achieves the highest HumanEval (77.44) among all external baselines at 25\%, outperforming RDS+ on both HumanEval and HumanEval+ while maintaining competitive MBPP performance.
\end{itemize}

\paragraph{Case Study: Defect Patterns in High RMI.}
To illustrate why extremely high RMI samples are harmful, we examine an example from the top RMI percentile (RMI $= 3.37$):

\begin{tcolorbox}[colback=blue!5!white, colframe=blue!50!white, boxrule=0.5pt]
\small
\textcolor{olive}{\textbf{Query}}: How to convert a \underline{gritty sulla matrix} with \underline{floating discharges} into a fairly \underline{crisp supra array} predominantly within \underline{specific thresholds} using Python?

\textcolor{teal}{\textbf{Answer}}: Converting a \underline{Gritty Sulla Matrix}... To convert a \underline{gritty sulla matrix} with \underline{floating discharges} into a fairly \underline{crisp supra array}... (followed by generic NumPy thresholding code)
\end{tcolorbox}

\noindent The query contains nonsensical jargon, and the answer simply paraphrases these terms (underlined). The answer makes the query highly predictable (hence high RMI) but provides little genuine learning signal.

\paragraph{Case Study: Low RMI as Misalignment.}
Conversely, we examine a sample with negative RMI (RMI $= -0.13$):

\begin{tcolorbox}[colback=blue!5!white, colframe=blue!50!white, boxrule=0.5pt]
\small
\textcolor{olive}{\textbf{Query}}: Hi!

\textcolor{teal}{\textbf{Answer}}: compensated, let's get you the help you need. Could you please provide more details about what you're working on or what specific Python coding question you have? I'm here to assist you with any Python-related challenges you might be facing. Whether it's a beginner question, a complex algorithm, or anything in between, feel free to share the details, and I'll do my best to help!
\end{tcolorbox}

\noindent The query ``Hi!'' is not a meaningful instruction for a code dataset. Consequently, the generated answer---despite attempting to be helpful---cannot provide information that helps predict the query. This is a typical example of \textbf{semantic misalignment}: a poorly-specified query leads to an answer that has little semantic connection to the query content (hence negative RMI). Such samples should be filtered out as they provide poor learning signal for code generation.

\paragraph{Math Domain Results.} Table~\ref{tab:math_results} reports results on MATH-500 and GPQA-Diamond. Using only 25\% of the data, QAQ approaches full-data performance on both benchmarks (91.6 vs.\ 90.6 on MATH-500; 39.9 vs.\ 42.4 on GPQA-D) while outperforming every other 25\%-data baseline, demonstrating that RMI generalizes beyond code generation. All baselines fall further below the full-data performance than QAQ on both benchmarks, confirming that naive data reduction without quality-aware selection is more harmful in the math domain.

\begin{table}[t]
\centering
\small
\begin{tabular}{lccc}
\toprule
\textbf{Method} & \textbf{Data Size} & \textbf{MATH-500} & \textbf{GPQA-D} \\
\midrule
Full Data$^\dagger$ & 100\% & 90.6 & 42.4 \\
\midrule
Random    & 25\% & 87.8 & 33.3 \\
IFD       & 25\% & 87.2 & 37.4 \\
SCAR      & 25\% & 86.6 & 30.8 \\
RDS+      & 25\% & 85.6 & 37.9 \\
\textbf{QAQ (Ours)} & \textbf{25\%} & \textbf{91.6} & \textbf{39.9} \\
\bottomrule
\end{tabular}
\caption{Results on math domain benchmarks. $^\dagger$Full Data is the official OpenR1-Qwen-7B checkpoint \citep{openr1} (Qwen2.5-Math-7B-Instruct base).}
\label{tab:math_results}
\end{table}

\subsection{Analysis}
\label{sec:analysis}

\paragraph{Why Does Model Disagreement Work?}
A single model's RMI conflates multiple factors that cannot be easily separated:
\begin{itemize}
    \item \textbf{Low RMI} may indicate: (1) the sample is too \textit{difficult} for the model to understand the Q-A relationship; (2) the Q-A pair is genuinely \textit{unrelated}; (3) the data is \textit{corrupted}; or (4) the model is already \textit{familiar} with this Q-A pattern, so the answer provides little additional information.
    \item \textbf{High RMI} may indicate: (1) the model recognizes a familiar pattern from pretraining; or (2) defect patterns such as paraphrasing or repetition.
\end{itemize}

By comparing models of different capabilities, we disentangle these signals: the strong model ensures \textit{validity} while the weak model ensures \textit{learnability}. Table~\ref{tab:disagreement} summarizes the four cases.

\begin{table}[t]
\centering
\small
\setlength{\tabcolsep}{3pt}
\begin{tabular}{cccm{3cm}}
\toprule
\textbf{Strong} & \textbf{Weak} & \textbf{Action} & \textbf{Interpretation} \\
\midrule
High & High & Filter & Defect patterns (e.g., repetition) \\
Low & Low & Filter & Corrupted, too hard, or trivial \\
Low & High & Filter & Easy for strong model \\
High & Low & \textbf{Retain} & Valid (strong) + learnable (weak) \\
\bottomrule
\end{tabular}
\caption{Model agreement/disagreement cases.}
\label{tab:disagreement}
\end{table}

This explains why Diff-based selection outperforms both single-model selection and consensus-based selection (Sum-High).

\paragraph{Ablation: Disagreement vs. Consensus.}
We compare four selection strategies under fixed hyperparameters (batch size 256, lr 4e-5, warmup 0.2); Table~\ref{tab:ablation} reports best-checkpoint results.

\begin{table}[t]
\centering
\small
\setlength{\tabcolsep}{3pt}
\begin{tabular}{lcccc}
\toprule
\textbf{Method} & \textbf{HumanEval} & \textbf{HumanEval+} & \textbf{MBPP} & \textbf{MBPP+} \\
\midrule
Diff-High & \textbf{77.44} & \textbf{71.95} & 71.43 & 58.73 \\
Sum-High & 74.39 & 68.90 & 71.16 & 59.52 \\
Sum-Low & 71.34 & 65.85 & 66.14 & 55.56 \\
Diff-Low & 71.34 & 67.07 & \textbf{74.87} & \textbf{62.43} \\
\bottomrule
\end{tabular}
\caption{Ablation on selection strategies (all 25\% data). Diff-High corresponds to QAQ (Ours) in Table~\ref{tab:main_results}.}
\label{tab:ablation}
\end{table}

Key observations: (1) \textbf{Diff $>$ Sum}: Diff-High outperforms Sum-High by 3 points on HumanEval+ (71.95 vs.\ 68.90), confirming disagreement is more informative than consensus. (2) \textbf{Sum-Low is worst}: Both-low agreement yields the poorest results, confirming such samples should be filtered. (3) \textbf{Diff-High vs Diff-Low}: Diff-High favors HumanEval while Diff-Low favors MBPP, suggesting the two signals capture complementary difficulty levels.

\paragraph{Stability of RMI Across Training.}
Pre- and post-finetuning RMI rankings correlate at $\rho = 0.9539$, confirming that RMI captures intrinsic data properties and is suitable for static one-time selection. The moderate cross-model correlation ($\rho = 0.70$) shows that RMI is also sensitive to model capability, which is precisely what enables the Diff signal. We discuss model-specific bias in Appendix~\ref{sec:appendix_bias}.

\paragraph{Sample Overlap Analysis.}
Diff-High and Sum-High select fundamentally different samples: their overlap is only \textbf{13.85\%}, explaining the 3-point HumanEval+ gap (71.95 vs.\ 68.90). Sum-High likely captures many paraphrasing cases---repetition is a simple pattern that both models detect, yielding high $r_s$ and $r_w$ simultaneously \citep{holtzman2020degeneration}---whereas Diff-High retains samples with genuine semantic content that only the strong model recognizes. Separately, strong and weak model top-50\% selections overlap at 76.1\%, yet yield different downstream performance, confirming that the 24\% non-overlapping samples carry disproportionate training signal.

We further verify generalization to Magpie-Qwen2.5-Coder-Pro-300K (Appendix~\ref{sec:appendix_magpie}), robustness to model pair choice (Appendix~\ref{sec:appendix_model_pair}), and stability across bin counts $K \in \{5,10,20,50\}$ (Appendix~\ref{sec:appendix_k_ablation}).

\section{Conclusion}

We proposed QAQ, a data selection framework that evaluates quality from the reverse direction $\text{PPL}(Q|A)$ and leverages model disagreement to identify samples that are both valid and learnable, demonstrating its effectiveness across code generation and math reasoning.

Our experiments reveal four key findings: (1) RMI-based selection with stratification consistently outperforms IFD-based selection, validating the importance of reverse direction analysis; (2) RMI outliers at both ends correlate with undesirable samples; (3) Model disagreement captures a unique subset of valuable samples (only 13.85\% overlap with consensus-based selection), yielding a 3-point HumanEval+ gain over consensus-based selection; (4) The method generalizes across domains: on math reasoning (OpenR1-Math-220k), QAQ approaches full-data performance with 25\% of the data while outperforming other 25\%-data baselines on both MATH-500 (91.6 vs.\ 90.6) and GPQA-Diamond (39.9 vs.\ 42.4), demonstrating that RMI is a quality signal that transfers across the two domains we tested (code generation and math reasoning).

Practically, selecting 25\% of data with QAQ achieves competitive performance on code generation and approaches full-data performance while outperforming other selection methods on math reasoning, offering a scalable path to reduce fine-tuning cost without sacrificing capability.

\section*{Limitations}

\paragraph{Dataset Diversity.} Main experiments use WarriorCoder, with validation on Magpie-Qwen2.5-Coder-Pro-300K (Appendix~\ref{sec:appendix_magpie}) and OpenR1-Math-220k (Section~\ref{sec:math}). Both code datasets are seedless synthetic and may share noise patterns; future work should cover seed-based corpora and general instruction-tuning datasets.

\paragraph{Model Selection.} The disagreement signal depends on the model pair. We use DeepSeek-Coder-6.7B-Base and Qwen3-0.6B; Appendix~\ref{sec:appendix_model_pair} shows robustness across three pairings including a non-code model, though the optimal pair for other domains or languages remains unexplored.

\paragraph{Hyperparameters.} We use $K=10$ as a default; ablations (Appendix~\ref{sec:appendix_k_ablation}) confirm stability across $K \in \{5,10,20,50\}$, suggesting the method is robust to this choice.

\paragraph{Reproducibility.} Results are single-run; future work should report multi-seed confidence intervals. Checkpoint selection is also not uniform across domains: training on the noisy WarriorCoder data is less stable across steps, so code-domain results select the best checkpoint, whereas math-domain training (following the OpenR1-Qwen-7B recipe) is more stable across steps, so we report the final training step for all methods.

\paragraph{Computational Cost.} RMI requires two forward passes per sample—a one-time overhead. See Appendix~\ref{sec:appendix_cost} for timing details.

\section*{Acknowledgments}

This work was supported by the Brain Science and Brain-like Intelligence Technology-National Science and Technology Major Project (grant no.\ 2021ZD0203702) to T.Y.

\bibliography{reference}

\clearpage
\appendix

\section{Generalization to Magpie-Qwen2.5-Coder-Pro-300K}
\label{sec:appendix_magpie}

To verify generalization beyond WarriorCoder, we evaluate on Magpie-Qwen2.5-Coder-Pro-300K \citep{magpie}, another seedless synthetic dataset. After filtering samples exceeding 2048 tokens, approximately 120K samples remain. We use the same model pair and select 25\% of data.

\begin{table}[h]
\centering
\small
\setlength{\tabcolsep}{1pt}
\begin{tabular}{lrcccc}
\toprule
\textbf{Method} & \textbf{\%} & \textbf{HumanEval} & \textbf{HumanEval+} & \textbf{MBPP} & \textbf{MBPP+} \\
\midrule
Full Data & 100 & 71.95 & 66.46 & 62.43 & 53.17 \\
\midrule
Random & 25 & 72.56 & 65.24 & 63.76 & 53.17 \\
IFD & 25 & 67.07 & 60.98 & 61.64 & 49.21 \\
SCAR & 25 & \textbf{73.78} & \textbf{66.46} & 63.23 & 49.74 \\
RDS+ & 25 & 70.73 & 64.02 & \textbf{69.58} & \textbf{57.67} \\
QAQ (Ours) & 25 & 71.95 & 65.24 & 68.25 & 56.35 \\
\bottomrule
\end{tabular}
\caption{Generalization to Magpie-Qwen2.5-Coder-Pro-300K. QAQ achieves the most balanced results across all four benchmarks; SCAR underperforms Full Data on MBPP+ (49.74 vs.\ 53.17), indicating benchmark-specific overfitting.}
\label{tab:generalization}
\end{table}

QAQ demonstrates competitive and balanced performance across all four benchmarks, without sacrificing any single metric. SCAR leads on HumanEval but drops below Full Data on MBPP+, suggesting benchmark-specific overfitting. RDS+ leads on MBPP/MBPP+ but underperforms on HumanEval+.

\section{Sensitivity to Model Pair}
\label{sec:appendix_model_pair}

To examine whether the cognitive gap signal depends on the specific model pair, we test two configurations beyond our original pairing: (1) DeepSeek-Coder-6.7B vs.\ Qwen2.5-Coder-7B (a stronger code model), and (2) DeepSeek-Coder-6.7B vs.\ Mistral-7B-v0.3 (a non-code-specialized model with a different architecture). All use Diff-High selection at 25\% data.

\begin{table}[h]
\centering
\scriptsize
\setlength{\tabcolsep}{2pt}
\begin{tabular}{lcccc}
\toprule
\textbf{Weak Model} & \textbf{HumanEval} & \textbf{HumanEval+} & \textbf{MBPP} & \textbf{MBPP+} \\
\midrule
Full Data (100\%) & \textbf{78.05} & 72.56 & 71.69 & 59.52 \\
\midrule
Qwen3-0.6B (orig.) & 77.44 & 71.95 & 71.43 & 58.73 \\
Qwen2.5-Coder-7B & 77.44 & \textbf{72.56} & 73.02 & 60.32 \\
Mistral-7B-v0.3 & 75.00 & 69.51 & \textbf{74.60} & \textbf{60.85} \\
\bottomrule
\end{tabular}
\caption{Sensitivity to model pair (strong model fixed as DeepSeek-Coder-6.7B-Base; 25\% data, Diff-High).}
\label{tab:model_pair}
\end{table}

Two findings stand out. First, Qwen2.5-Coder-7B is a stronger code model than DeepSeek-Coder-6.7B, yet pairing them matches or exceeds Full Data on three of four metrics (HumanEval+, MBPP, MBPP+)---showing that the method works even when the designated ``weak'' model is actually stronger. Second, Mistral-7B-v0.3, a non-code-specialized model with a different architecture, still yields effective selection. Together, these results indicate that the key factor is \textit{meaningful disagreement} between the two models, rather than a strict capacity gap.

\section{Sensitivity to Binning Parameter $K$}
\label{sec:appendix_k_ablation}

The number of bins $K$ primarily affects mid-range selection: a coarser partition mixes samples across a wider complexity range within each bin, while a finer partition enforces stricter complexity matching but reduces within-bin sample size. Table~\ref{tab:k_ablation} shows that performance is stable across $K \in \{5, 10, 20, 50\}$, with $K = 5$--$10$ performing best overall, confirming the method is robust to this hyperparameter.

\begin{table}[h]
\centering
\small
\setlength{\tabcolsep}{3pt}
\begin{tabular}{lcccc}
\toprule
$K$ & \textbf{HumanEval} & \textbf{HumanEval+} & \textbf{MBPP} & \textbf{MBPP+} \\
\midrule
5  & \textbf{78.66} & 70.73 & \textbf{74.07} & \textbf{60.58} \\
10 & 77.44 & \textbf{72.56} & 71.43 & 58.47 \\
20 & 74.39 & 70.73 & 73.28 & 60.58 \\
50 & 75.00 & 70.73 & 73.28 & 60.05 \\
\bottomrule
\end{tabular}
\caption{Ablation on number of PPL($Q$) bins $K$ (RMI 50-75\%, 25\% data).}
\label{tab:k_ablation}
\end{table}

\section{Implementation Details and Model Configurations}
\label{sec:appendix_prompts}

To provide full transparency on how Reverse Mutual Information (RMI) is calculated, we present the exact chat templates and system prompts used for both the strong and weak models.

\paragraph{DeepSeek-Coder Template} 
The following template follows the official convention for DeepSeek-Coder-Instruct.

\begin{tcolorbox}[colback=blue!5!white, colframe=blue!50!white, boxrule=0.5pt, title=System Template for DeepSeek-Coder]
\small
\textbf{System}: You are an AI programming assistant, utilizing the Deepseek Coder model, developed by Deepseek Company, and you only answer questions related to computer science. For politically sensitive questions, security and privacy issues, and other non-computer science questions, you will refuse to answer. \\

\end{tcolorbox}

\paragraph{Qwen3 Template} 
For Qwen3-0.6B, we use a consistent system prompt to focus the model on computer science tasks. Note that for this model, we explicitly set \texttt{enable\_thinking=False} to maintain standard autoregressive behavior.

\begin{tcolorbox}[colback=blue!5!white, colframe=blue!50!white, boxrule=0.5pt, title=System Template for Qwen3]
\small
\textbf{System}: You are an AI programming assistant, and you only answer questions related to computer science. For politically sensitive questions, security and privacy issues, and other non-computer science questions, you will refuse to answer. 
\end{tcolorbox}

\paragraph{Implementation Note} In both cases, $\text{PPL}(Q|A)$ is calculated by performing a forward pass on the ground-truth question tokens represented by \texttt{[Q]} in the Assistant turn, while teacher-forcing the System and User inputs as context.

\section{Sensitivity to Chat Template and Task Framing}
\label{sec:appendix_template}

We investigate how specific task instructions and formatting affect RMI and IFD signals. 

\paragraph{Standard Implementations (Template-based).}
In our main experiments, both metrics utilize structured chat templates. For **Standard RMI**, the reverse generation task is highly constrained: \texttt{``TASK: Given an answer, generate the most likely \textbf{computer science} question that this answer is responding to. If the inferred question is outside computer science, respond with \textbf{`INVALID'}.''} 

For **Template-based IFD**, we compute the perplexity of the answer $A$ within the following structured templates:

\begin{tcolorbox}[colback=blue!5!white, colframe=blue!50!white, boxrule=0.5pt]
\small
\textbf{System}: Standard system prompt \\
\textbf{User}: [None] \\
\textbf{Assistant}: \texttt{[A]}
\end{tcolorbox}

\begin{tcolorbox}[colback=blue!5!white, colframe=blue!50!white, boxrule=0.5pt]
\small
\textbf{System}: Standard system prompt \\
\textbf{User}: \texttt{[Q]} \\
\textbf{Assistant}: \texttt{[A]}
\end{tcolorbox}

\paragraph{Variations for Sensitivity Analysis.}
To isolate the impact of task framing, we compare our standard method against several variations:
\begin{itemize}
    %%% NEW CONTENT: RAW IFD DEFINITION %%%
    \item \textbf{Raw IFD}: Following the original implementation proposed by \citet{cherry_llm}, Raw IFD is computed via simple text concatenation: \texttt{[Q] + [A]} for the conditional part, and \texttt{[A]} for the unconditional part, without any system prompts or role indicators (User/Assistant).
    %%% END NEW CONTENT %%%
    \item \textbf{Simplified Task Prompt ($\rho=0.88$)}: Removes both the domain specificity and the negative constraint, using a more general instruction: \texttt{``TASK: Given an answer, generate the most likely question that this answer is responding to.''}
    %%% ADDED THIS ITEM %%%
    \item \textbf{System Context Only ($\rho=0.79$)}: For $PPL(Q)$, we retain the System Prompt but \textbf{remove the explicit ``TASK'' instruction}. This tests whether the model's persona alone (e.g., ``You are an AI assistant'') is sufficient for estimating query probability.
    
    \item \textbf{Prefix-only ($\rho=0.70$)}: For $PPL(Q)$, we strip away the System Prompt and use only a minimal newline-delimited indicator (\texttt{"Question:"}).
    
    \item \textbf{Pure Raw ($\rho=0.62$)}: For $PPL(Q)$, we calculate the perplexity from the very first token of the query with \textbf{zero context} (cold start). 
\end{itemize}

\paragraph{Results and Analysis.} 
As shown in Table~\ref{tab:template_sensitivity}, IFD is remarkably stable ($\rho=0.957$) even under raw concatenation. In contrast, RMI drops as we strip away the template and context. The correlation decreases from 0.88 (Simplified Template) to 0.79 (System-Prefix), 0.70 (Prefix-only), and 0.62 (Pure Raw). This confirms that while the specific \textit{wording} of the template matters less, the presence of \textbf{sufficient context} for $PPL(Q)$ is critical. Even without a task template, providing the system prompt ($\rho=0.79$) significantly stabilizes the metric compared to raw inputs.

\begin{table}[h]
\centering
\small
\begin{tabular}{lc}
\toprule
\textbf{Comparison (vs. Standard RMI)} & \textbf{Spearman $\rho$} \\
\midrule
IFD vs. Raw IFD & 0.957 \\
RMI vs. Simplified Task Prompt & 0.88 \\
RMI vs. System-Prefix (No Template) & 0.79 \\
RMI vs. Prefix-only (No Template) & 0.70 \\
RMI vs. Pure Raw (No Template) & 0.62 \\
\bottomrule
\end{tabular}
\caption{Spearman correlation of metrics under different task framings.}
\label{tab:template_sensitivity}
\end{table}

\begin{figure}[ht]
\centering
\includegraphics[width=0.8\columnwidth]{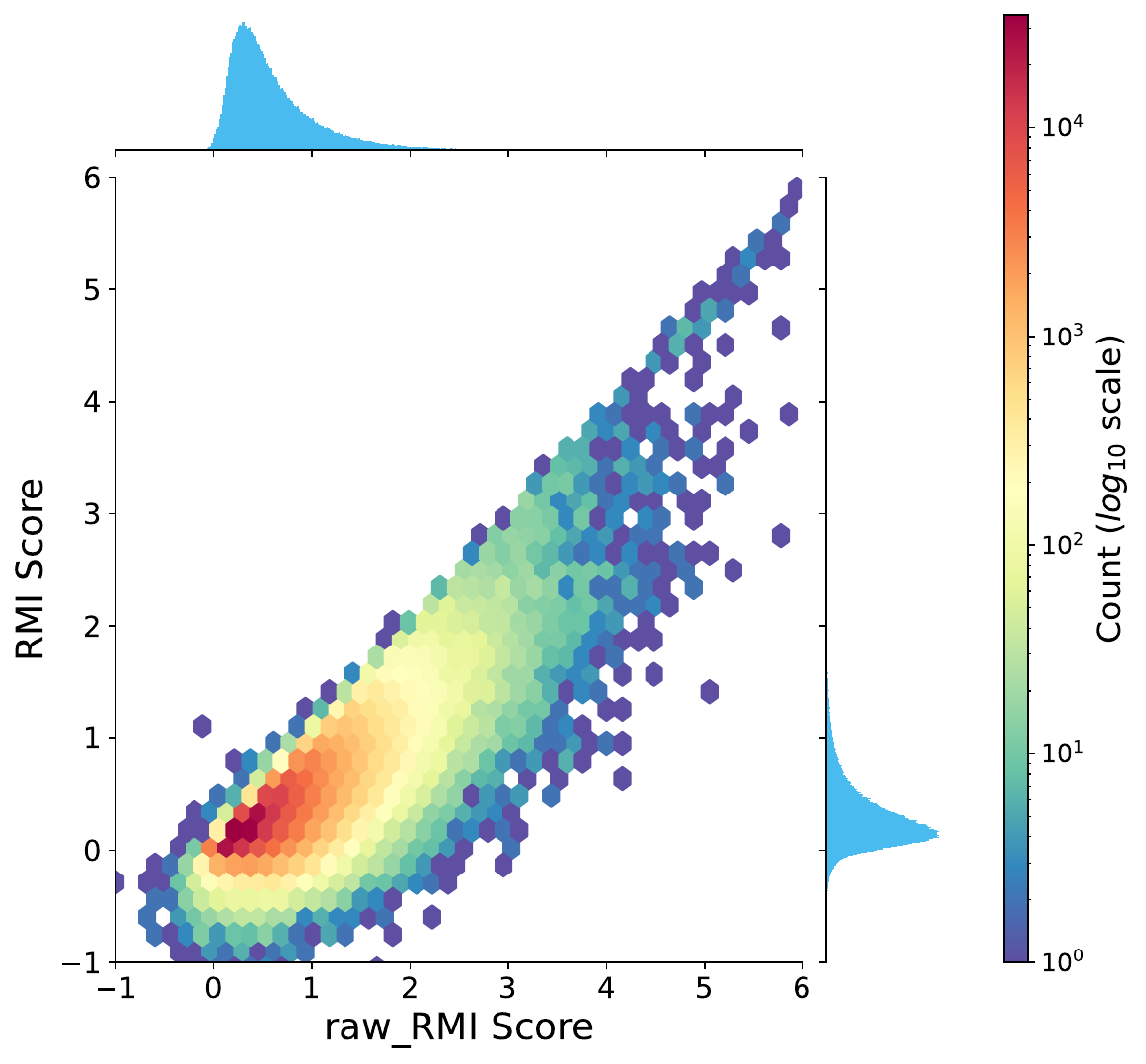}
\caption{Scatter plot comparing Standard RMI vs. Raw RMI. The lack of system context in the Raw version leads to inflated $PPL(Q)$ and subsequent overestimation of RMI.}
\label{fig:raw_vs_standard_rmi}
\end{figure}

The primary source of instability is the \textbf{contextual imbalance in $PPL(Q)$ calculation}. In the \textit{Pure Raw} setting, the model must predict the query from a "cold start" without knowing it is handling a CS query. Since queries are typically short, the high loss incurred by the initial tokens dominates the average $PPL(Q)$, biasing the $RMI$ upwards and introducing noise (see Figure~\ref{fig:raw_vs_standard_rmi}).

\paragraph{Discussion: Why IFD is More Robust.}
IFD remains robust because $PPL(A)$ is calculated on answer sequences (code) that are typically longer and follow familiar structural patterns (\texttt{def}, \texttt{import}). Moreover, $A$ is always conditioned on $Q$ in the IFD calculation, avoiding the "zero context" problem. The correlation jump from 0.62 to 0.70 by simply adding a \texttt{"Question:"} indicator proves that task framing is essential for RMI to capture genuine semantic coherence rather than prefix-induced surprise.

Table~\ref{tab:template_downstream} further confirms this: Standard RMI 50-75\% outperforms Simplified RMI by 2--3 points on HumanEval(+), and both RMI variants substantially outperform IFD even when IFD uses an enhanced semantic-constrained prompt. This confirms that the core effectiveness of QAQ comes from the reverse direction ($A \rightarrow Q$) rather than prompt engineering.

\begin{table}[h]
\centering
\scriptsize
\setlength{\tabcolsep}{3pt}
\begin{tabular}{lcccc}
\toprule
\textbf{Method} & \textbf{HumanEval} & \textbf{HumanEval+} & \textbf{MBPP} & \textbf{MBPP+} \\
\midrule
Standard RMI 50-75\%   & \textbf{77.44} & \textbf{72.56} & \textbf{71.43} & \textbf{58.47} \\
Simplified RMI 50-75\% & 75.00 & 69.51 & 70.37 & 58.20 \\
IFD w/ enhanced prompt & 70.12 & 65.85 & 65.08 & 55.82 \\
\bottomrule
\end{tabular}
\caption{Downstream performance under different template variants (25\% data, RMI 50-75\%; HE = HumanEval).}
\label{tab:template_downstream}
\end{table}

\section{On Model-Specific Bias and Self-Reinforcement}
\label{sec:appendix_bias}

RMI scores are inherently model-dependent: different models produce somewhat different rankings, with cross-model Spearman correlations ranging from 0.63 to 0.80. We acknowledge this model-specific bias exists.

However, this is distinct from \textit{self-reinforcement}, which would require a feedback loop where training amplifies the selection bias. This does not occur for three reasons. First, training optimizes $\text{PPL}(A|Q)$, not RMI, which is computed via $P(Q|A)$---the two objectives are in opposite directions. Second, RMI is a static, one-time metric computed before training begins and is never updated. Third, empirically, RMI rankings remain stable after fine-tuning (Spearman $\rho > 0.9$, see Section~\ref{sec:analysis}), confirming no systematic shift.

The model-specific bias is therefore static (originating from pretraining) and does not amplify during fine-tuning. Involving the target model's perspective in data selection is a reasonable design choice, analogous to curriculum learning, where training materials are tailored to the learner's current knowledge state.

\section{Computational Cost}
\label{sec:appendix_cost}

Table~\ref{tab:cost} reports wall-clock time for computing selection scores on the WarriorCoder dataset (310K samples) using 8$\times$A100 GPUs.

\begin{table}[h]
\centering
\small
\begin{tabular}{lll}
\toprule
\textbf{Method} & \textbf{Time} & \textbf{Notes} \\
\midrule
SCAR         & $\sim$1.5h & Small embedding model \\
RDS+         & $\sim$2h   & Single pass on Q and A \\
RMI (single) & $\sim$3h   & PPL over Q only \\
IFD          & $\sim$6h   & Two passes over A \\
QAQ (Ours)   & $\sim$6h   & Two RMI computations \\
\bottomrule
\end{tabular}
\caption{Wall-clock time for score computation on WarriorCoder 310K (8$\times$A100 GPUs).}
\label{tab:cost}
\end{table}

RMI single-pass is faster than IFD because it operates over questions (Q), which are typically shorter than answers (A) in code datasets. The full QAQ pipeline (Cognitive Gap) requires two model inferences, bringing total selection time to $\sim$6h.

\end{document}